\documentclass[runningheads]{llncs}
\usepackage[T1]{fontenc}
\usepackage{comment}
\usepackage{amsmath,amssymb}
\usepackage{hyperref}
\usepackage{bm}
\usepackage{multirow}
\usepackage{booktabs}   
\usepackage{graphicx,verbatim}
\usepackage{color}

\begin{document}
\title{Spinverse: Differentiable Physics for Permeability-Aware Microstructure Reconstruction from Diffusion MRI}
\titlerunning{Spinverse}
%



\author{Prathamesh Pradeep Khole\inst{1}\thanks{Equal contribution} \and
Mario M. Brenes\inst{1}\textsuperscript{$\star$} \and
Zahra Kais Petiwala\inst{2} \and
Ehsan Mirafzali\inst{1} \and
Utkarsh Gupta\inst{1} \and
Jing-Rebecca Li\inst{3} \and
Andrada Ianus\inst{4} \and
Razvan Marinescu\inst{1}}
\authorrunning{P.P. Khole, M.M. Brenes et al.}
\institute{University of California Santa Cruz, Santa Cruz, CA, USA\\
\email{pkhole@ucsc.edu} \and
University of California San Diego, San Diego, CA, USA \and
Inria-Saclay, France \and
Kings College London, London, UK}

  
\maketitle              
\begin{abstract}
Diffusion MRI (dMRI) is sensitive to microstructural barriers, yet most existing methods either assume impermeable boundaries or estimate voxel-level parameters without recovering explicit interfaces. We present \textit{Spinverse}, a permeability-aware reconstruction method that inverts dMRI measurements through a fully differentiable Bloch--Torrey simulator. Spinverse represents tissue on a fixed tetrahedral grid and treats each interior face permeability as a learnable parameter; low-permeability faces act as diffusion barriers, so microstructural boundaries whose topology is not fixed a priori (up to the resolution of the ambient mesh) emerge without changing mesh connectivity or vertex positions. Given a target signal, we optimize face permeabilities by backpropagating a signal-matching loss through the PDE forward model, and recover an interface by thresholding the learned permeability field. To mitigate the ill-posedness of permeability inversion, we use mesh-based geometric priors; to avoid local minima, we use a staged multi-sequence optimization curriculum. Across a collection of synthetic voxel meshes, Spinverse reconstructs diverse geometries and demonstrates that sequence scheduling and regularization are critical to avoid outline-only solutions while improving both boundary accuracy and structural validity.
\end{abstract}

\section{Introduction}

Diffusion MRI (dMRI) is sensitive to micron-scale barriers that restrict
diffusion and enable exchange, offering a noninvasive window into tissue
microstructure \cite{alexander2019microstructure,novikov2019quantifying}. Yet, \emph{reconstructing explicit microstructural boundaries}
from the measured signal remains challenging: the signal is indirect, integrates
over sub-voxel geometries, and multiple configurations can produce similar
attenuation profiles. Most approaches therefore estimate \emph{voxel-level}
parameters via simplified compartment models such as
NODDI~\cite{zhang2012noddi} and SANDI~\cite{palombo2020sandi}, which provide
millimeter-scale biomarkers but do not recover \emph{where} barriers are, and
typically treat permeability as negligible.

Physics-based simulators~\cite{li2019spindoctor} and simulation-driven
pipelines~\cite{fang2023simulation} offer high-fidelity forward prediction in
complex geometries, yet their outputs are signals or aggregate parameter
estimates rather than explicit barrier reconstructions. 
Recent differentiable solvers~\cite{khole2024ddmrisim,khole2025remidireconstructionmicrostructureusing} enable inverse mesh reconstruction from dMRI but remain ill-posed without strong priors; those works optimized vertex coordinates in a low-dimensional auto-encoded spectral space under an arbitrary but fixed topology.
A method that optimizes meshes of variable topologies
would be a significant step towards the reconstruction of accurate diffusion MRI
geometries.

In this work, we introduce Spinverse, a method that reconstructs
microstructural boundaries whose topology is not fixed a priori
(up to the resolution of the ambient mesh) by optimizing face-wise
permeabilities on a fixed tetrahedral grid. Our contributions are: \textbf{(1)} we formulate
\textit{permeability-aware boundary reconstruction} from dMRI as an inverse
problem over a \textit{face-wise permeability field} on a volumetric mesh, where
boundaries emerge as low-permeability interfaces; \textbf{(2)} we implement an
\textit{end-to-end differentiable} Bloch--Torrey~\cite{PhysRev.104.563} simulator
that treats each tetrahedron as an independent compartment with face
permeabilities as optimizable parameters. Building on
\cite{li2019spindoctor,khole2025remidireconstructionmicrostructureusing,khole2024ddmrisim},
we re-formulate the forward model in PyTorch~\cite{paszke2019pytorchimperativestylehighperformance}
with parallelized per-element solves, enabling scalable gradient-based inversion
via Autograd; and \textbf{(3)} we propose an inversion strategy combining
\textit{spatial priors} (MRF-based continuity~\cite{Zhang2001,Boykov2006,Paulsen2010}
and manifold regularization~\cite{Belkin2004}) with a \textit{multi-sequence
curriculum} to mitigate ill-posedness and improve interior boundary recovery.

\section{Related Work}
Compartment models (e.g., NODDI~\cite{zhang2012noddi}, SANDI~\cite{palombo2020sandi})
estimate voxel-level parameters but do not recover explicit barriers or mesh
permeabilities. FEM Bloch--Torrey simulators such as SpinDoctor~\cite{li2019spindoctor} and \cite{khole2024ddmrisim}
predict signals in complex geometries with permeable interfaces, and
simulation-driven learning maps signals to aggregate parameters~\cite{fang2023simulation}. Differentiable inverse solvers such as
ReMiDi~\cite{khole2025remidireconstructionmicrostructureusing} reconstruct mesh
geometry under a Bloch--Torrey \cite{PhysRev.104.563} forward model but assume
fixed mesh topology; Spinverse instead infers boundaries by optimizing a face-wise
permeability field on a fixed mesh grid, with interfaces emerging via the coupling
operator $\mathbf{B}(\kappa)$ in~\eqref{eq:semidiscrete}, so the extracted
interface topology is determined entirely by the optimization and is not
constrained to match a predefined template, unlike vertex-based approaches that
inherit the topology of their initial mesh. More broadly, differentiable
mesh/topology relaxations such as
DMesh++~\cite{son2025dmeshefficientdifferentiablemesh} motivate soft-to-hard
optimization, which we adopt under a Bloch--Torrey PDE \cite{PhysRev.104.563}
based forward model.

\begin{figure}[t]
\centering
\includegraphics[width=0.9\textwidth, trim=0 0 0 0, clip]{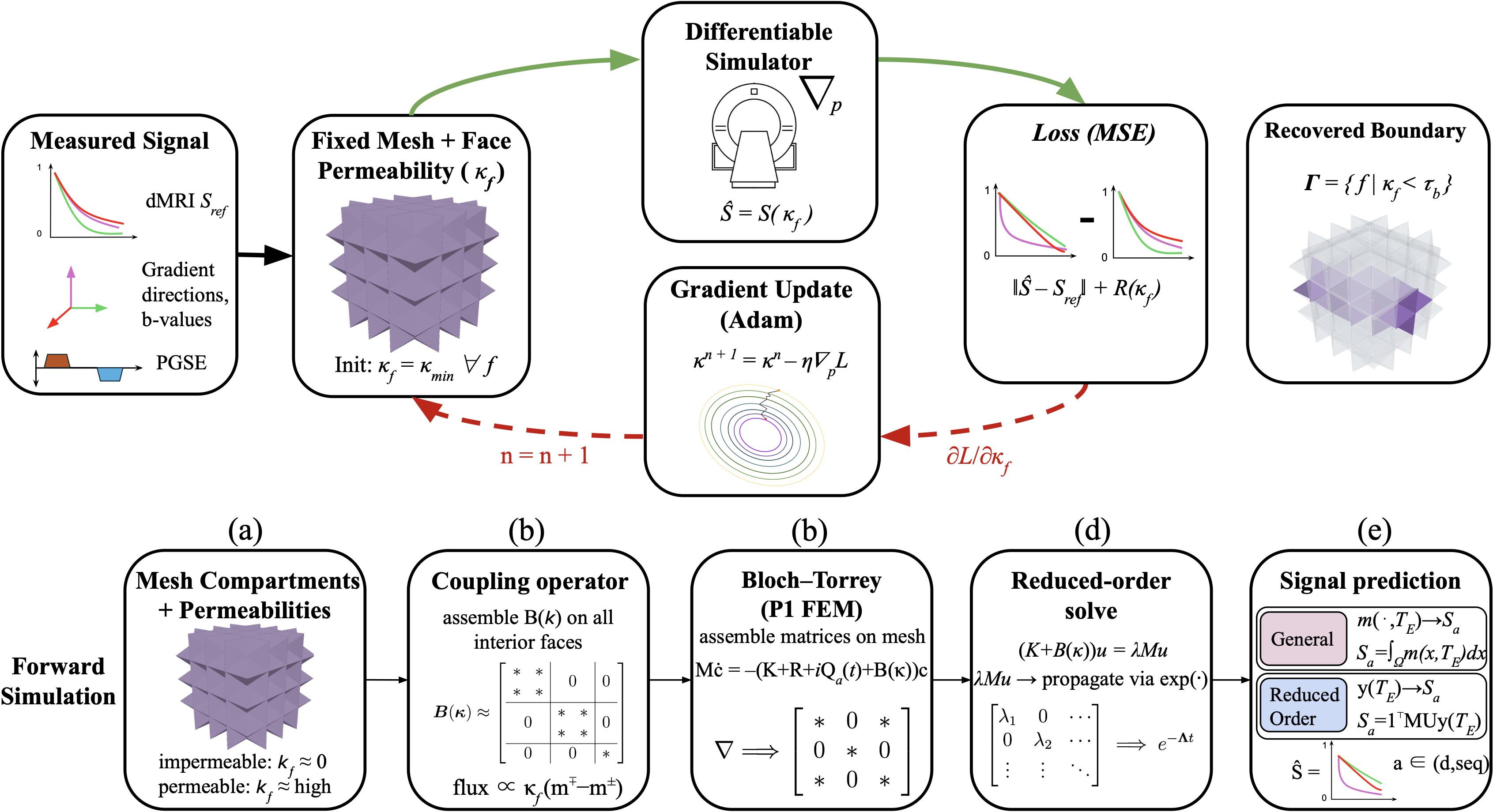}

\caption{\textbf{Spinverse pipeline.}
\emph{Top:} Given reference dMRI signals $S_{\mathrm{ref}}$ under multiple PGSE encodings, we optimize face permeabilities $\{\kappa_f\}$ on a fixed tetrahedral mesh by minimizing MSE loss with regularization. Gradients are backpropagated through a differentiable Bloch--Torrey simulator and used by Adam to update $\kappa_f$, recovering boundaries via thresholding.
\emph{Bottom (a--e):} Simulator internals: compartments and face permeabilities~(a); coupling operator $\mathbf{B}(\kappa)$ assembly~(b); Bloch--Torrey FEM system~(c); reduced-order eigensolve and matrix-exponential propagation~(d); signal readout~(e).}

\label{fig:pipeline}
\end{figure}

\section{Methods}

\textbf{Problem setup: per-element compartments and learnable permeabilities}. \label{sec:method_setup} Let $\mathcal{M}=(\mathcal{V},\mathcal{T})$ be an ambient tetrahedral mesh
discretizing a bounded domain $\Omega\subset\mathbb{R}^3$.
Rather than pre-segmenting $\mathcal{M}$ into tissue compartments, we treat
\emph{every tetrahedron as an independent compartment}.
Each interior face $f\in\mathcal{F}_{\mathrm{int}}$ (shared by exactly two
tetrahedra) becomes a semi-permeable interface with learnable permeability
$\kappa_f>0$ (Fig.~\ref{fig:pipeline}a).
We optimize unconstrained parameters $\theta_f\in\mathbb{R}$ and obtain physical permeabilities via a tempered sigmoid reparametrization in log-space,
$k_f = \log_{10}\kappa_f = k_{\min} + (k_{\max}-k_{\min})\,\sigma(\theta_f/\tau_\sigma)$,
where $\sigma$ is the logistic sigmoid, $\tau_\sigma>0$ a temperature, and $[k_{\min},k_{\max}]$ bounds the log-permeability range.
Vertex positions and mesh connectivity are fixed; microstructure emerges
solely through $\{\kappa_f\}$, and a barrier set is recovered post-hoc as
$\Gamma(\tau_b)=\{f\in\mathcal{F}_{\mathrm{int}}:\kappa_f<\tau_b\}$
(Fig.~\ref{fig:pipeline}, top).


\noindent \textbf{Differentiable forward model}. \label{sec:method_forward}
The transverse magnetization $m(\mathbf{x},t)$ at position
$\mathbf{x}\in\Omega$, time $t\in[0,T_E]$, satisfies the Bloch--Torrey
equation~\cite{li2019spindoctor} under diffusion encoding $a\in\mathcal{A}$:
\begin{equation}
\frac{\partial m}{\partial t}
=\nabla\!\cdot\!(\mathbf{D}\nabla m)-\frac{1}{T_2}\,m
-i\gamma\,\mathbf{G}_a(t)\!\cdot\!\mathbf{x}\,m,
\quad m(\mathbf{x},0)=\rho,
\label{eq:bt}
\end{equation}
where $\mathbf{D}$ is the diffusion tensor, $T_2$ the transverse relaxation time, $\gamma$ the gyromagnetic ratio, $i$ the imaginary unit, $\rho$ the proton density, and $\mathbf{G}_a(t)$ the PGSE gradient waveform of acquisition $a$ (encoding direction, $\delta$, $\Delta$, $b$-value).
Under per-element compartmentalization, every interior face carries the
interface condition
$\mathbf{n}^{\pm}\!\cdot\!\mathbf{D}^{\pm}\nabla m^{\pm}
=\kappa_f(m^{\mp}-m^{\pm})$,
where superscripts $\pm$ denote the two tetrahedra sharing face $f$ and
$\mathbf{n}^{\pm}$ their outward normals.
This reduces to a reflecting barrier as $\kappa_f\!\to\!0$
(Fig.~\ref{fig:pipeline}a).
The predicted signal is $S_a(\kappa)=\int_\Omega m(\mathbf{x},T_E)\,d\mathbf{x}$.

Discretizing~\eqref{eq:bt} with P1 (piecewise-linear) finite elements
following~\cite{li2019spindoctor} yields the semi-discrete system (Fig.~\ref{fig:pipeline}b--c):
\begin{equation}
\mathbf{M}\dot{\mathbf{c}}
=-\bigl(\mathbf{K}+\mathbf{R}+i\mathbf{Q}_a(t)
+\mathbf{B}(\kappa)\bigr)\mathbf{c},
\label{eq:semidiscrete}
\end{equation}
where $\mathbf{M},\mathbf{K},\mathbf{R},\mathbf{Q}_a$ are mass,
stiffness, relaxation, and gradient dephasing matrices.
The coupling operator $\mathbf{B}(\kappa)$ is assembled on \emph{all}
interior faces from face-area-weighted Robin coupling terms; it depends
linearly on $\{\kappa_f\}$ and is the \emph{only} operator updated during
optimization.
For efficiency we project into a truncated generalized eigenbasis of
$(\mathbf{K}+\mathbf{B}(\kappa))\mathbf{u}=\lambda\mathbf{M}\mathbf{u}$
and propagate in the reduced coordinates via matrix exponentials
(Fig.~\ref{fig:pipeline}d--e)~\cite{li2019spindoctor,fang2023simulation,khole2025remidireconstructionmicrostructureusing},
reading out the signal as
$S_a=\mathbf{1}^\top\mathbf{M}\mathbf{U}\mathbf{y}(T_E)$.
In practice, we refresh the truncated basis every $N$ iterations and reuse it between refreshes; on intermediate iterations we update only the reduced interface operator $\mathbf{U}^\top\mathbf{B}(\kappa)\mathbf{U}$ while treating $\mathbf{U}$ as fixed, so gradients flow through the reduced projections and time propagation back to $\{\theta_f\}$.

\noindent \textbf{Optimization: data fidelity, priors, and curriculum}. \label{sec:method_opt}
We estimate $\theta^\star$ by minimizing (Fig.~\ref{fig:pipeline}, top)
\begin{equation}
\theta^\star
=\arg\min_{\theta}
\sum_{a\in\mathcal{A}(t)}
\lambda_{data}\mathcal{L}_{data}\!\bigl(S_a(\kappa(\theta)),S_a^\star\bigr)
+\lambda_{\mathrm{cont}}\,\mathcal{R}_{\mathrm{cont}}\!\bigl(\kappa(\theta)\bigr)
+\lambda_{\mathrm{man}}\,\mathcal{R}_{\mathrm{man}}\!\bigl(\kappa(\theta)\bigr)
\label{eq:full_objective}
\end{equation}
where $\kappa(\theta)=\{\kappa_f(\theta)\}_{f\in\mathcal{F}_{\mathrm{int}}}$ are face permeabilities obtained from unconstrained parameters $\theta$ via a positive mapping (Sec.~\ref{sec:method_setup}). We use a weighted squared error on normalized signals,
$\mathcal{L}_{\mathrm{data}}(S_a,S_a^\star)=\|S_a/\widehat{S}_0 - S_a^\star/S_0^\star\|_2^2$,
where $\widehat{S}_0$ and $S_0^\star$ are predicted and measured $b{=}0$ references.

The face-wise field admits many signal-equivalent configurations, so we regularize using mesh adjacency.
\emph{Weighted continuity} penalizes discrepancies between adjacent faces while allowing sharp transitions at \emph{candidate interfaces} by downweighting face pairs likely to straddle a boundary. Specifically, $\mathcal{R}_{\mathrm{cont}}(\kappa)=\sum_{(f,f')\in\mathcal{N}} w_{ff'}\,(\kappa_f-\kappa_{f'})^2$,
with weights $w_{ff'}=1-|p_{\mathrm{non}}(f)-p_{\mathrm{non}}(f')|$.
Here $\mathcal{N}$ is the set of adjacent interior-face pairs and $p_{\mathrm{non}}(f)$ is a soft indicator of a non-permeable (barrier) face computed from $\kappa_f$.

We introduce a \emph{soft manifold regularizer} that penalizes mesh edges 
with other than $0$ or $2$ incident boundary faces, discouraging 
non-manifold junctions.
For an edge $e$ with incident faces $\mathcal{F}_e$, let $k_e=\sum_{f\in\mathcal{F}_e} p_{\mathrm{non}}(f)$; we penalize deviations from $\{0,2\}$ via
$\mathcal{R}_{\mathrm{man}}(\kappa)=\sum_{e\in\mathcal{E}} \operatorname{softmin}(k_e^2,\,(k_e-2)^2)$, where $\operatorname{softmin}(a,b)=$ $-\tau_m\log(e^{-a/\tau_m}+e^{-b/\tau_m})$ with temperature $\tau_m>0$.

Diffusion acquisitions probe complementary physical regimes: long diffusion times allow spins to encounter barriers and exchange across interfaces, making the signal sensitive to the coarse compartment layout whereas short diffusion times probe local restriction and fine boundary details.
Na\"{\i}vely optimizing all regimes jointly can induce gradient conflict.
We therefore use a staged curriculum: we first optimize on long diffusion-time acquisitions (exchange-dominated) to establish a coarse interface configuration, then transition to short diffusion-time acquisitions (restriction-dominated) to refine fine-scale structure.
Formally, we partition acquisitions into $\mathcal{A}_{\mathrm{long}}$ and $\mathcal{A}_{\mathrm{short}}$ and set
$\mathcal{A}(t)=\mathcal{A}_{\mathrm{long}}$ for $0\le t<T_{\mathrm{switch}}$ and $\mathcal{A}(t)=\mathcal{A}_{\mathrm{short}}$ for $T_{\mathrm{switch}}\le t\le T$,
where $T$ is the total number of iterations and $T_{\mathrm{switch}}$ is fixed per experiment (Sec.~\ref{sec:experiments}).

\section{Experiments and Results}
\label{sec:experiments}
We evaluate reconstructions with metrics for signal fidelity, geometric accuracy, and surface validity.
The section is organized as qualitative reconstructions, ablations of priors and scheduling, and a comparison to learned predictors on multi-compartment axon-like scenes.

\begin{figure}[t]
\centering
\begin{minipage}{0.38\textwidth}
  \centering
  \includegraphics[width=\linewidth]{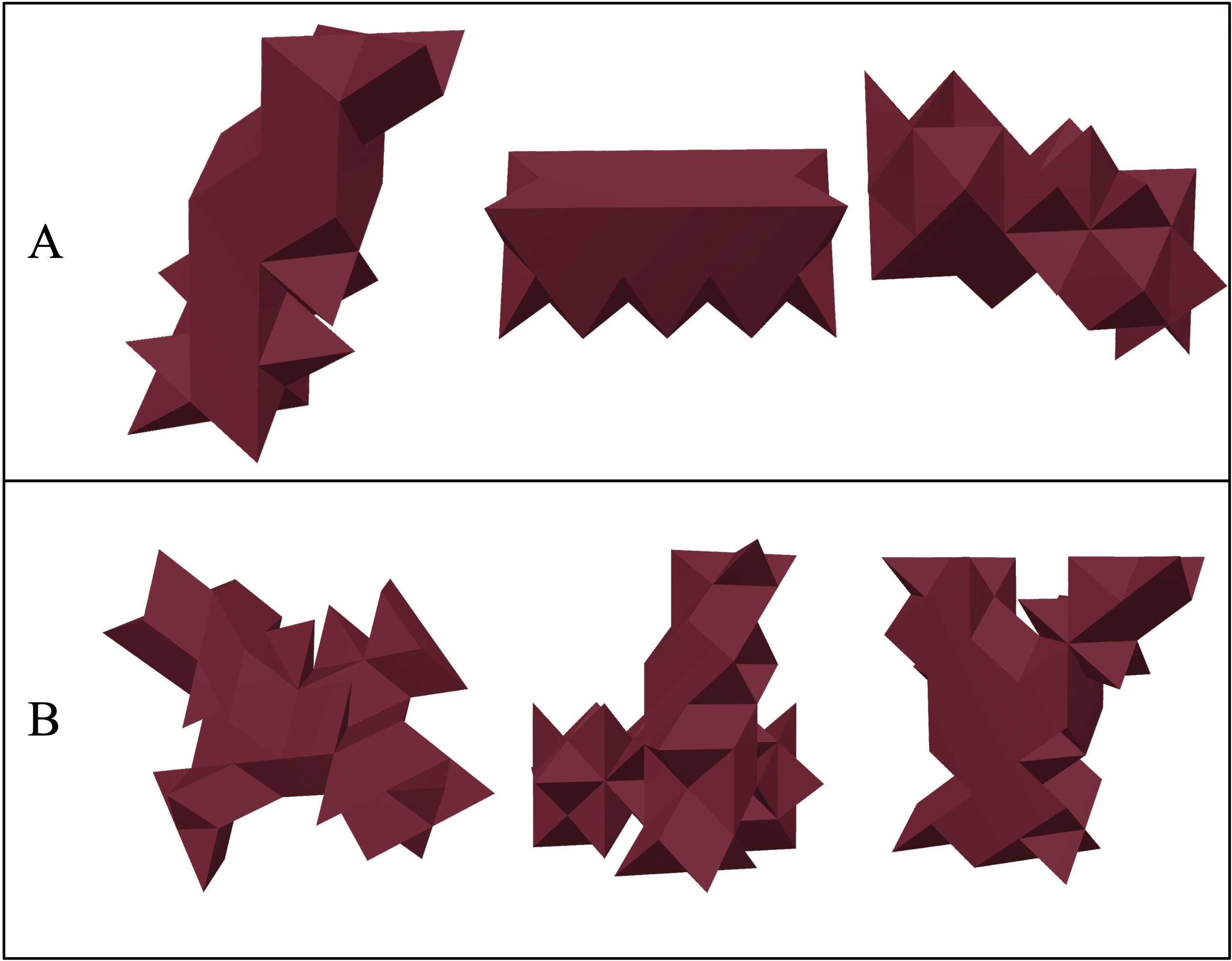}
\end{minipage}\hfill
\begin{minipage}{0.525\textwidth}
  \centering
  \includegraphics[width=\linewidth]{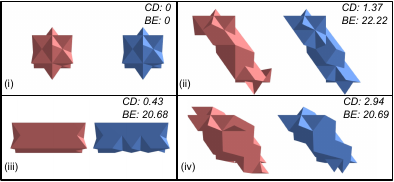}
\end{minipage}
\caption{\textbf{Sample data and reconstructions.}
(Left)~Representative ground-truth meshes: single-axon~(A) and two-axon~(B) geometries embedded in the ambient tetrahedral grid.
(Right) Four example Spinverse reconstructions (red) shown alongside ground truth meshes (blue). Each pair reports CD-L2 (CD) and BadEdge\% (BE).}
\label{fig:data_and_recons}
\end{figure}

\noindent \textbf{Setup.}
We evaluate our method on synthetic voxel-scale meshes of varying
geometry (e.g., sphere, axons, torus) on a fixed cubic domain
$x,y,z\in[-13.6,13.6]\;\mu\mathrm{m}$.
Signals are simulated under PGSE with $b\in\{0,1000,2000,3000,4000,5000\}$ s/mm$^2$, $\delta=10$ ms, and $\Delta\in\{20,60\}$ ms (short/long diffusion time).
We optimize a face-wise permeability field $\kappa_f\in[10^{-5},10^{-1}]$ and extract the interface by thresholding at $\tau_b=10^{-3}$.
Reconstructions are initialized at the threshold ($\kappa_f=10^{-3}$), while reference meshes assign low permeability ($\kappa_f=10^{-5}$) to boundary faces (all other faces set to $\kappa_f=10^{-1}$). In our protocol, $\mathcal{A}_{\mathrm{short}}$ denotes acquisitions with $\Delta=20$ ms and
$\mathcal{A}_{\mathrm{long}}$ denotes acquisitions with $\Delta=60$ ms (Sec.~\ref{sec:method_opt}).
We optimize $\theta$ with Adam \cite{DBLP:journals/corr/abs-1711-05101,kingma2017adammethodstochasticoptimization} (base learning rate $\eta_0=0.75$) using a cyclic warmup+cosine schedule: each $C=200$-iteration cycle linearly warms up for $W=50$ iterations, then cosine-decays to $\alpha\eta_0$ with $\alpha=0.1$. We set $\lambda_{\mathrm{data}}=100$,  $\lambda_{\mathrm{cont}}=2$, and linearly ramp $\lambda_{\mathrm{man}}$ from $0$ to $2.0$ over the first $400$ iterations. We run for 400 iterations and switch acquisitions at $T_{\mathrm{switch}}=200$.

\noindent \textbf{Metrics.}
We measure (i) \emph{signal fidelity} using MSE on normalized signals $S/S_0$, averaged across all $b$-values and gradient directions; (ii) \emph{geometric accuracy} using symmetric Chamfer distance with squared Euclidean distances (CD-L2) \cite{DBLP:journals/corr/FanSG16}, computed after centering each point set sampled from boundary faces (consistent with translation invariance of the diffusion MRI signal); and (iii) \emph{structural validity} using edge-incidence statistics. Let $d(e)$ be the number of boundary faces incident to edge $e$ (restricted to edges with $d(e)>0$); we report BadEdge\% (BE\%) ($d(e)\neq 2$) to quantify deviations from a two-manifold interface.

\begin{figure}[t]
\centering
\includegraphics[width=1.0\textwidth, trim=0 0 0 0, clip]{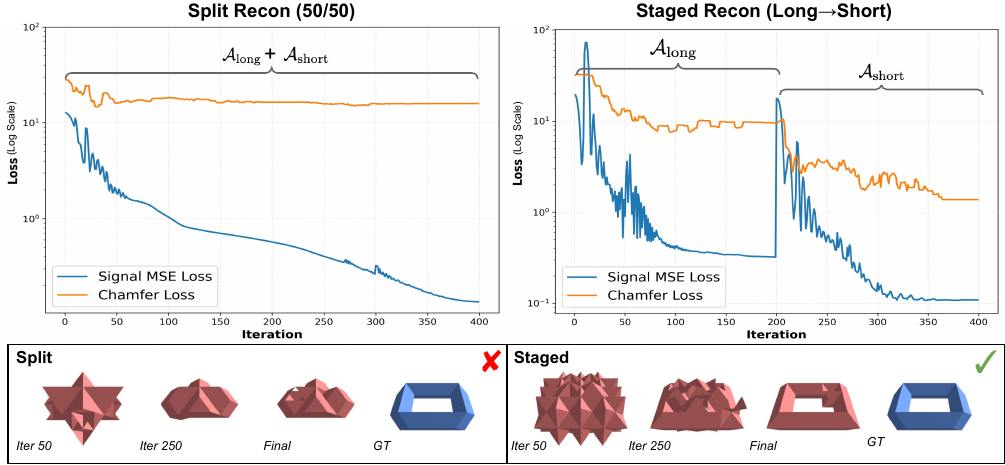}
\caption{\textbf{Acquisition scheduling ablation (torus).} Joint 50/50 optimization over $\mathcal{A}_{\mathrm{long}}\cup\mathcal{A}_{\mathrm{short}}$ (\emph{Split}) versus a staged schedule $\mathcal{A}_{\mathrm{long}}\!\rightarrow\!\mathcal{A}_{\mathrm{short}}$ (\emph{Staged}). Top: signal MSE and Chamfer (CD-L2) over iterations. Bottom: extracted interfaces at iterations 50/250/final (optimized, red) and ground truth (blue); staging yields a coherent torus boundary while joint optimization fails to recover the opening.}
\label{fig:schedule_ablation}
\end{figure}

\noindent \textbf{Qualitative reconstructions.}
Fig.~\ref{fig:data_and_recons}(i)–(iv) shows representative final reconstructions across geometries (optimized interface in red, ground truth in blue).
Spinverse accurately recovers the sphere (i) and diagonal cylinder (ii) from a barrier-free initialization (CD-L2 $0.00$/$1.37$, BE\% $0.00$/$20.15$), and the additional examples in (iii)–(iv) exhibit similar cylindrical boundary recovery with comparable structural artifacts from thresholding. See the supplementary material for videos of these reconstructions, including the torus case.

\noindent \textbf{Acquisition schedule ablation (torus).}
We focus on the torus as a challenging case with nontrivial topology and an interior opening (Fig.~\ref{fig:schedule_ablation}).
Joint 50/50 optimization over $\mathcal{A}_{\mathrm{long}}\cup\mathcal{A}_{\mathrm{short}}$ reduces signal MSE but plateaus in Chamfer and converges to a fragmented interface that fails to recover the opening.
In contrast, a staged schedule $\mathcal{A}_{\mathrm{long}}\!\rightarrow\!\mathcal{A}_{\mathrm{short}}$ progressively refines the boundary (iter~50 $\rightarrow$ 250 $\rightarrow$ final), yielding the correct topology (CD-L2 $1.37$, BE\% $22.78$).
This ablation highlights an optimization failure mode when combining acquisition regimes that probe complementary diffusion physics: jointly fitting all regimes can produce gradient conflict and lead to local minima. Staging instead provides a coarse-to-fine path that first localizes barriers under long diffusion time and then sharpens geometry under short diffusion time.

\begin{figure}[t]
\centering
\begin{minipage}[c]{0.48\textwidth}
    \centering
    \includegraphics[width=\linewidth]{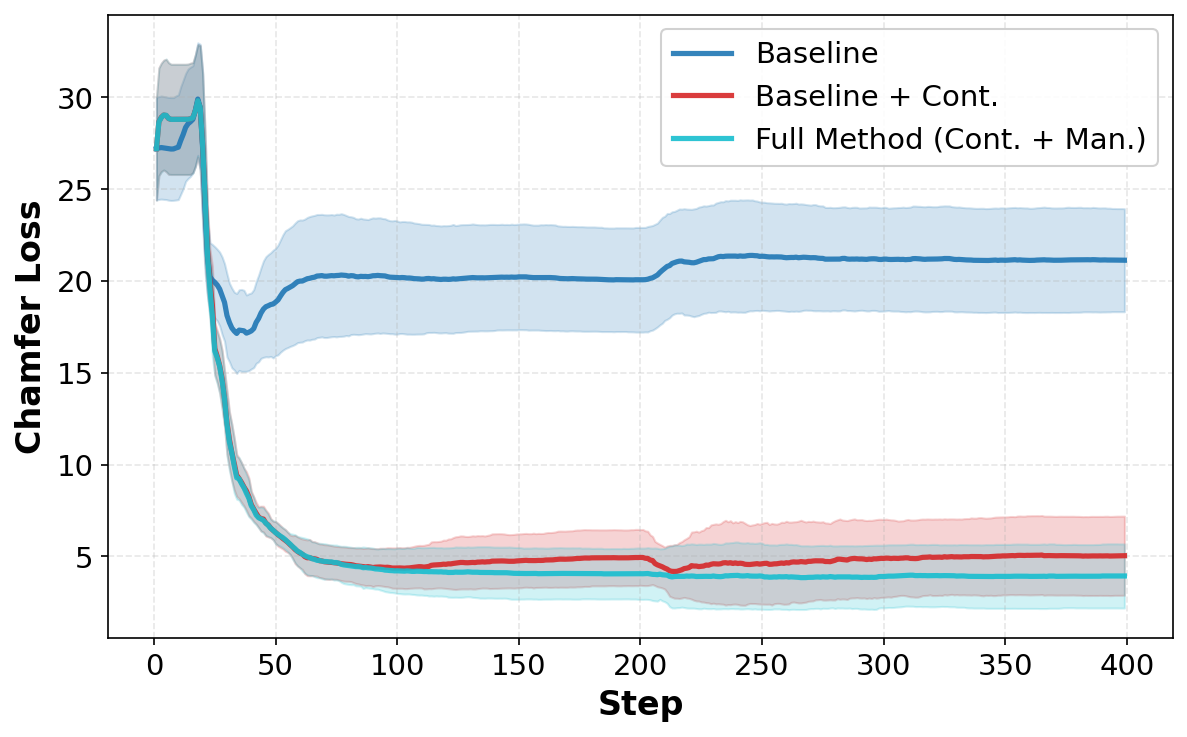}
\end{minipage}
\hfill
\begin{minipage}[c]{0.48\textwidth}
    \centering
    \small
    \setlength{\tabcolsep}{4pt}
    \begin{tabular}{lcc}
        \toprule
        \textbf{Config} & \textbf{CD-L2} $\downarrow$ & \textbf{BE\%} $\downarrow$ \\
        \midrule
        Baseline     & $21.1 \pm 2.8$ & $66.3 \pm 4.8$\\
        + Cont.      & $5.0 \pm 2.2$  & $45.4 \pm 8.8$\\
        + Cont./Man. & $\mathbf{3.9 \pm 1.7}$ & $\mathbf{27.5 \pm 7.7}$\\
        \bottomrule
    \end{tabular}
\end{minipage}
\caption{The left-hand plot shows average CD-L2 loss over 10 meshes for (i) baseline method (no regularization), (ii) baseline with continuity regularization and (iii) the full method with continuity/manifold regularization. The table shows final chamfer distance and BadEdge\% values for each test configuration.}
\label{fig:ablation_plot_table}
\end{figure}

\noindent \textbf{Ablations on regularization.}
Fig.~\ref{fig:ablation_plot_table} shows that adding continuity regularization substantially improves geometric recovery, reducing CD-L2 from $21.1\pm2.8$ to $5.0\pm2.2$ and decreasing BE\% from $66.3\pm4.8$ to $45.4\pm8.8$.
Adding the manifold regularizer further improves interface structure, reducing BE\% to $27.5\pm7.7$ while also improving CD-L2 to $3.9\pm1.7$. Overall, continuity primarily drives the large gain in geometric accuracy, while the manifold term provides the largest additional reduction in non-manifold artifacts without sacrificing CD-L2. In absolute terms, BE\% remains non-zero (e.g., $\approx 27.5\%$), reflecting that the thresholded face set can form open or branching interface complexes. In practice we observed a tradeoff: overly strong manifold regularization can suppress necessary boundary changes and degrade geometric recovery, whereas the chosen weight provides a favorable balance, improving interface validity without increasing CD-L2.

\begin{figure}[t]
\centering
\includegraphics[width=0.65\textwidth, trim=0 0 0 0, clip]{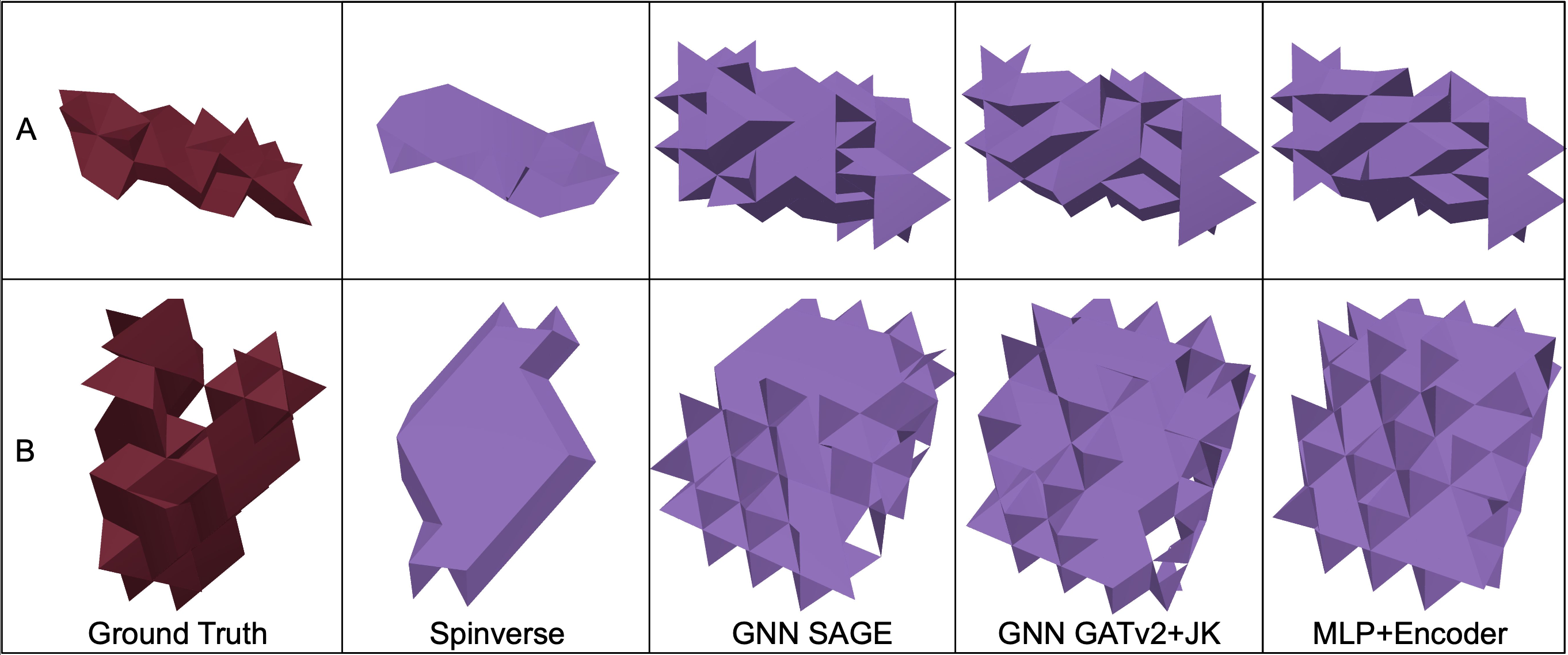}
\caption{\textbf{Comparison with learned predictors} on single-axon~(A) and two-axon~(B) meshes. Spinverse recovers compact boundaries; baselines produce many spurious faces.}
\label{fig:nn_comparison}
\end{figure}

\begin{table}[t]
\caption{CD-L2 and BE\% (mean $\pm$ std) across 100 test reconstructions each.}
\label{tab:chamfer_combined}
\centering
\small
\setlength{\tabcolsep}{3pt}
\begin{tabular}{|l|l|c|c|c|c|}
\hline
\textbf{Geometry} & \textbf{Metric} & \textbf{MLP} & \textbf{GraphSAGE} & \textbf{GATv2+JK} & \textbf{Spinverse} \\
\hline
\multirow{2}{*}{1 Axon}
  & CD-L2 $\downarrow$  & $10.81{\pm}5.95$ & $10.75{\pm}6.26$ & $8.91{\pm}4.25$ & $\mathbf{4.19{\pm}1.57}$ \\
  & BE\% $\downarrow$ & $67.12{\pm}6.47$ & $68.18{\pm}8.56$ & $71.07{\pm}8.91$ & $\mathbf{23.40{\pm}9.10}$ \\
\hline
\multirow{2}{*}{2 Axon}
  & CD-L2 $\downarrow$  & $\mathbf{8.28{\pm}2.42}$ & $9.03{\pm}2.63$ & $8.42{\pm}2.52$ & $9.79{\pm}3.64$ \\
  & BE\% $\downarrow$ & $73.81{\pm}4.25$ & $75.76{\pm}3.65$ & $72.77{\pm}5.82$ & $\mathbf{35.62{\pm}5.64}$ \\
\hline
\end{tabular}
\end{table}

\noindent \textbf{Comparison to learned predictors}.
We compare Spinverse against amortized baselines: an MLP~\cite{rumelhart1986backprop} with learned encodings and two GNN variants (GraphSAGE~\cite{NIPS2017_5dd9db5e} and GATv2~\cite{DBLP:journals/corr/abs-2105-14491} with jumping knowledge~\cite{DBLP:journals/corr/abs-1806-03536}) that predict face-level barrier probabilities in a single forward pass. All networks are trained with Focal Loss~\cite{DBLP:journals/corr/abs-1708-02002} and FiLM conditioning~\cite{DBLP:journals/corr/abs-1709-07871} on 1000 synthetic bent-cylinder meshes (single-axon and two-axon crossings, Fig.~\ref{fig:data_and_recons}) and evaluated on 100 held-out meshes; Spinverse runs 400 iterations of per-instance optimization without any training data.
Table~\ref{tab:chamfer_combined} reports CD-L2 and BE\%.
On single-axon meshes, Spinverse roughly halves CD-L2 ($4.19$ vs.\ $8.91$) and reduces BE\% by ${\sim}3\times$ ($23.4\%$ vs.\ $67\text{--}71\%$). On the harder two-axon setting, baselines achieve marginally lower
CD-L2 ($8.28\text{--}9.03$ vs.\ $9.79$) but at the cost of
${\sim}2\times$ higher non-manifold artifacts (BE\%
$72\text{--}76\%$ vs.\ $35.6\%$); the CD-L2 advantage is consistent
with over-prediction of barrier faces, which improves point-cloud
coverage but degrades structural validity
(Fig.~\ref{fig:nn_comparison}B).
Fig.~\ref{fig:nn_comparison} confirms this qualitatively: baselines exhibit spurious faces and fragmented structure, whereas Spinverse recovers coherent boundaries. 

\noindent \textbf{Discussion.}
Spinverse performs most reliably on single-compartment geometries, where the long$\rightarrow$short diffusion-time curriculum supports a coarse-to-fine refinement, recovering the dominant boundary structure before sharpening fine-scale barrier details. 
Continuity and manifold priors further regularize the ill-posed inverse problem, improving both geometric accuracy (CD-L2) and interface validity (BE\%). BE\% remains non-zero in several cases, and additional post-processing or stronger topology constraints may be required. We find that multi-compartment meshes remain challenging: 
reconstructions collapse toward symmetric envelopes that 
exploit the rotational degeneracy of the diffusion signal.
This regime is inherently more ambiguous, as a larger set of interfaces must be inferred from the same limited acquisition set, exacerbating non-identifiability.
Moreover, regularization strengths selected for single-compartment geometries may not transfer; stronger continuity/manifold priors may over-simplify the interface configuration. 
In terms of compute, each reconstruction takes ${\sim}33$\ minutes on a single 24GB NVIDIA L4 GPU and the solver's $O(n^2)$ memory scaling per element currently limits the mesh resolution that can be handled. 
Finally, experiments are limited to controlled synthetic voxels with known forward-model parameters, and robustness to noise remains to be evaluated.

\section{Conclusion}
We presented Spinverse, a physics-informed method that reconstructs
microstructural boundaries from diffusion MRI by optimizing face-wise
permeabilities on a fixed tetrahedral mesh through a differentiable
Bloch--Torrey simulator. Our experiments show that mesh-based geometric
priors and staged multi-sequence curricula are critical for recovering
coherent, structurally valid interfaces, and that the physics-based
inversion produces cleaner boundaries than learned baselines without
requiring training data. Current evaluation is limited to synthetic
geometries, and improving Spinverse on multi-compartment settings
remains an open challenge. Future work will explore hybrid learned initialization with physics-informed refinement, more scalable solvers to enable finer mesh resolutions, task-driven sequence design that optimizes acquisition schedules to probe specific microstructural features, and validation on experimentally acquired dMRI data.

\begin{credits}
\subsubsection{\ackname}
This work used computing resources provided by the National Research Platform (NRP) Nautilus cluster. We thank the NRP team for their support.
\end{credits}

\bibliographystyle{splncs04}
\bibliography{MICCAI}

\end{document}